\documentclass[conference]{IEEEtran}
\IEEEoverridecommandlockouts
\usepackage{cite}
\usepackage{url}
\usepackage{amsmath,amssymb,amsfonts,bm}
\usepackage[ruled,norelsize]{algorithm2e}
\ifCLASSOPTIONcompsoc
  \usepackage[caption=false, font=normalsize, labelfont=sf, textfont=sf]{subfig}
\else
  \usepackage[caption=false, font=footnotesize]{subfig}
\fi

\usepackage{graphicx}
\usepackage{textcomp}
\usepackage{todonotes}
\usepackage{xcolor}
\def\BibTeX{{\rm B\kern-.05em{\sc i\kern-.025em b}\kern-.08em
    T\kern-.1667em\lower.7ex\hbox{E}\kern-.125emX}}
    
\begin{document}
\title{Layer-Parallel Training with GPU Concurrency of Deep Residual Neural Networks via Nonlinear Multigrid\\
\thanks{Research was sponsored by the United States Air Force Research Laboratory and was accomplished under Cooperative Agreement Number FA8750-19-2-1000. The views and conclusions contained in this document are those of the authors and should not be interpreted as representing the official policies, either expressed or implied, of the United States Air Force or the U.S. Government. The U.S. Government is authorized to reproduce and distribute reprints for Government purposes notwithstanding any copyright notation herein.}
}

\author{
\IEEEauthorblockN{Andrew Kirby, Siddharth Samsi, Michael Jones, Albert Reuther, Jeremy Kepner, Vijay Gadepally}
\IEEEauthorblockA{\textit{Massachusetts Institute of Technology Lincoln Laboratory Supercomputing Center} \\
Lexington, MA, USA}
}

\maketitle

\begin{abstract}
A Multigrid Full Approximation Storage algorithm for solving Deep Residual Networks is developed to enable neural network parallelized layer-wise training and concurrent computational kernel execution on GPUs. This work demonstrates a 10.2x speedup over traditional layer-wise model parallelism techniques using the same number of compute units. 
\end{abstract}

 
\section{Introduction}
Improvements in computing power have driven resurgence and significant advancements in Machine Learning and, specifically, Deep Learning \cite{Hoefler:2019,Reuther:2019}. To realize these improvements, the transition from the Dennard-scaling era of increasing CPU clock frequencies through the miniaturization of transistors to the Multicore era \cite{Leiserson:2020} has compelled parallel algorithmic development. For Supervised Learning, the primary parallel methodology employed is data parallelism which trains neural network models by evaluating multiple data samples simultaneously on separate hardware components, computing their respective model parameter gradients with respect to the input data sample, and collectively averaging the gradients which are then used to update the model parameters.

However, in the post-Moore's-Law era composed of increasingly parallel hardware (within and between compute units), data parallelism suffers from global collective communication patterns, which limit the long-term performance capabilities. Further, data parallelism does not address model size limitations or sequential evaluation of the neural network using forward or backward propagation. An orthogonal approach, which can be paired with data parallelism, to address  model-size limitation is model parallelism, which partitions a model across multiple compute units. However, traditional model parallelism approaches also fail to address the sequential evaluation of neural networks, thereby being constrained in performance by Amdalh's Law\footnote{Amdalh's Law states the maximum speedup of an algorithm is limited by the fraction of sequential procedure in a program.}. To circumvent Amdahl's Law, we develop an asynchronous layer-parallel algorithm for Graphical Processing Units (GPU) enabling constant asymptotic execution time independent of the number of neural network layers. 

\begin{figure*}[t]
\centerline{\includegraphics[width=0.8\linewidth]{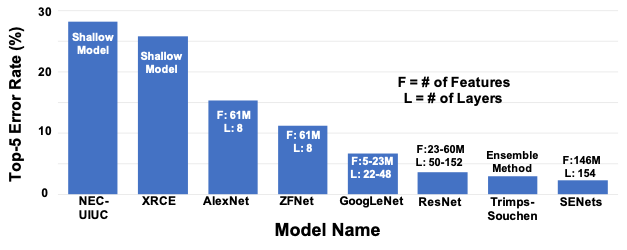}}
\caption{Analysis of ILSVRC Winners from 2010-2017. The y-axis corresponds to Top-5 error rate (\%) of winning entries. An observable trend in image classification and other applications is the correlation between performance and increased network depth.}
\label{fig:imagenetdepth}
\end{figure*}

Beyond hardware trends, a number of applications such as image classification, video recognition, natural language processing have relied on increasingly deep neural networks. For example, Figure~\ref{fig:imagenetdepth} describes winning entries from the ImageNet Large Scale Visual Recognition Challenge (ILSVRC) \cite{Russakovsky:2015}. In this and other similar applications, deeper networks can often provide greater performance (e.g., accuracy). However, training such networks requires significant increase in computational power required. For example, the authors of \cite{Amodei:2018} indicate a 3.4 month doubling time in computing requirements. A significant number of deep neural network architectures fall into the class of deep residual networks.

\section{Deep Residual Networks} \label{ResNets}
Deep Residual Networks have been demonstrated to have superior performance for supervised learning techniques applied to real-world applications such as image classification problems \cite{Kaiming:2016} and various other technologies \cite{Gadepally:2019,Samsi:2019}, and have resolved network training issues due to \textit{vanishing gradients} by enacting a skip connection allowing gradient information to accumulate and bypass multiplicative operators.

Residual neural networks (ResNet) are composed of residual blocks stacked in $N$-layers of the following structure:
\begin{equation}
\bm{u}^{n+1} = \bm{u}^{n} + hF\left(\bm{u}^{n};\bm{\theta}^{n}\right),\ \ \text{for } n = 0, \ldots, N - 1 \label{EQ:resnet}
\end{equation}
with $\bm{u},\bm{\theta} \in \mathbb{R}^{q}$ being the layer state and parameters, respectively, and $q$ is the layer width, $F$ is the feature transformation operator (e.g. activation of convolution and bias), and $h$ is a scaling factor.

A ResNet can be naturally described mathematically as an Initial Value Problem (IVP) such that 
\begin{equation}
\frac{\partial \bm{u}(t)}{\partial t} = F\left(\bm{u}(t); \bm{\theta}(t)\right),\ \ \forall t \in T \label{EQ:IVP}
\end{equation} 
where $T \subseteq \mathbb{R}^{+} \cup \left\{0\right\}$. Equation \eqref{EQ:IVP} can be discretized by the explicit Forward Euler Method \cite{Yu:2017,Chen:2018} leading to the formulation in \eqref{EQ:resnet} where $h$ represents the discrete time step size. 

\section{Multigrid Methods}
Multigrid (MG) methods are iterative algorithms that accelerate the solution convergence of a system of equations.
The technique is traditionally applied in scientific and engineering applications such as problems found in Computational Fluid Dynamics \cite{Mavriplis:1990,Mavriplis:1995}, and, more recently, in Deep Learning including the acceleration of training optimization \cite{Merzhauser:2020} and layer parallelism on CPUs \cite{Gunther:2020}. This work follows \cite{Gunther:2020} and extends the algorithm to GPUs and enables concurrent kernel execution through NVIDIA's CUDA Streams \cite{Nickolls:2008} and Deep Neural Network (CuDNN) \cite{Chetlur:2014}.

\subsection{Multigrid Principles}
The principle idea of Multigrid is to use multiple \textit{levels} of resolution to iteratively solve an error prescribed by the residual equation governed by the system. Suppose we wish to solve for the unknown $u$ in the following:
\begin{IEEEeqnarray}{rCl}
Lu &=& f \label{EQ:continuous}\\
r &:=& f - L\overline{u} \label{EQ:residual}
\end{IEEEeqnarray}
$L$ is a functional (linear or nonlinear) operating on the solution $u$ which is equal to a known source $f$. Given an approximate solution $\overline{u}$, the residual $r$ is defined by \eqref{EQ:residual}; note that if $\overline{u} = u$, then $r = 0$. 
The construction of the MG algorithm is independent of the set of equations or discretization setting. These principles will be used to demonstrate the construction of multigrid algorithm for solving the neural network states via forward propagation. Consider the solution of the discrete problem:
\begin{equation}
L_{h}u_{h} = f_{h} \label{EQ:exact_h}
\end{equation}
where the subscripts refer to the discretization of the continuous formulation \eqref{EQ:continuous} on a \textit{level} of step size $h$.
Let $\overline{u}_h$ be the current estimate of $u_h$ obtained by approximately solving \eqref{EQ:continuous}. Since $\overline{u}_h$ does not satisfy the equation exactly, then the non-zero residual $r_h$ is given as:
\begin{equation}
r_{h} = f_{h} - L_{h}\overline{u}_{h} \label{EQ:resid_h}
\end{equation}
The objective of Multigrid is to solve for the solution correction $\delta u_h$ such that the exact solution is given by:
\begin{equation}
u_{h} = \overline{u}_{h} + \delta u_{h} \label{EQ:delta_u}
\end{equation}
Subtracting \eqref{EQ:exact_h} from \eqref{EQ:resid_h} yields the following:
\begin{equation}
L_{h}u_{h} - L_{h}\overline{u}_{h} = r_{h} \label{EQ:correction}
\end{equation}
If $L_{h}$ is a linear operator, then substituting \eqref{EQ:delta_u} into \eqref{EQ:correction} results in the correction equation:
\begin{equation}
L_{h}\delta u_{h} = r_{h} \label{EQ:linear_correction}
\end{equation}
Assuming the errors on level $h$ are high-frequency, the correction $\delta u_h$ can be computed more efficiently on a coarser level $H$ by solving the equation
\begin{equation}
L_{H} \delta u_{H} = I_{h}^{H} r_{h} \label{EQ:linear_resid}
\end{equation}
where $I_{h}^{H}$ denotes a restriction operator which interpolates the residual from the fine level $h$ to the coarse level $H$. If \eqref{EQ:linear_resid} contains low computational work, it may be solved exactly by direct or iterative methods. Alternatively, the above procedure may be applied recursively on coarser levels to yield an approximate solution to the correction $\delta u_{H}$ itself. Once the correction is computed, an update is employed on the fine level to correct the solution:
\begin{equation}
\overline{u}^{new}_{h} = \overline{u}^{old}_{h} + I^{h}_{H}\delta u_{H}
\end{equation}
where $I^{h}_{H}$ is a prolongation operator which interpolates the coarse level correction to the fine level. 
Following the fine-level update, the solution may be solved approximately again. This process is known as relaxation which may be repeated until an acceptable level of error tolerance is achieved. Note that when the fine level solution is achieved, then $r_h = 0$ and \eqref{EQ:linear_correction} is solved trivially $\delta u_H = 0$. Thus no further corrections to the fine level solution are produced.

\subsection{Full Approximation Storage (FAS)}
Suppose the operator in \eqref{EQ:correction} is nonlinear, then the equation can no longer be replaced by $L_h \delta u_h$, thus modifications are needed to accommodate this nonlinear behavior. This is done by introducing a new coarse level variable $\overline{u}_{H}$:
\begin{equation}
\overline{u}_{H} = \overline{I}_{h}^{H} \overline{u}_{h} + \delta u_{H} \label{EQ:coarse_grid_update}
\end{equation}
where $\overline{I}_{h}^{H}$ represents the restriction operator which transfers the fine level solution to the coarse level.
The analogue nonlinear equation to \eqref{EQ:linear_resid} is as follows:
\begin{equation}
L_{H}\overline{u}_{H} - L_{H}\overline{I}_{h}^{H} \overline{u}_{h} = I_{h}^{H} r_{h} \label{EQ:nonliner_resid}
\end{equation}
Recall the middle and right terms of \eqref{EQ:nonliner_resid} are known quantities leaving $\overline{u}_H$ as the only unknown variable.
Thus it is useful to introduce a new variable $S_H$ such that
\begin{IEEEeqnarray}{rCl}
L_{H}\overline{u}_{H} = S_H \label{EQ:coarse_equivalent} \\
\text{where}\ \ \ S_H := L_{H}\overline{I}_{h}^{H} \overline{u}_{h} + I_{h}^{H} r_{h}
\end{IEEEeqnarray}
On the coarse level, the set of equations defined by \eqref{EQ:coarse_equivalent} are analogous to \eqref{EQ:exact_h} so similar solution techniques can be applied to solve for the coarse level correction. Lastly, once the coarse level correction $\overline{u}_H^{new}$ is solved by \eqref{EQ:coarse_equivalent}, the fine level solution variables are updated as follows:
\begin{equation}
\overline{u}^{new}_{h} = \overline{u}^{old}_{h} + I^{h}_{H}\left( \overline{u}_H^{new} - \overline{I}_h^H \overline{u}_h^{old} \right)
\end{equation}
which, utilizing \eqref{EQ:coarse_grid_update}, is written finally as
\begin{equation}
\overline{u}^{new}_{h} = \overline{u}^{old}_{h} + I^{h}_{H}\delta u_H
\end{equation}

\begin{figure*}[t]
\centerline{\includegraphics[width=0.7\linewidth]{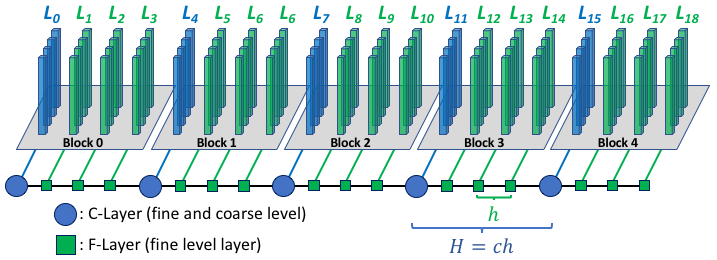}}
\caption{Neural network layer partitioning with coarsening factor $c=4$. Each layer block is composed of a $c$ layers.}
\label{fig:coarsening-layers}
\end{figure*}

\subsection{Multigrid for Residual Network Forward Propagation}
Residual Networks correspond to a time-discretized Ordinary Differential Equation IVP, and our goal is to remove the serial propagation of information by forward and backward propagation through network layers. We apply the iterative FAS MG scheme to the layers of the network. The application of Multigrid to the time dimension in Computational Physics is known as the Multigrid Reduction In Time (MGRIT) algorithm \cite{Falgout:2014,Dobrev:2017} and has been demonstrated to achieve significant speedups when sufficient computational resources are utilized \cite{Falgout:2017a,Falgout:2017b,Gunther:2019}. This algorithm allows the layers to be partitioned into local blocks, which can be solved concurrently and in parallel, thus, breaking Amdahl's Law limitation within the training process of neural networks. We will cast Residual Networks as the nonlinear system in \eqref{EQ:continuous} to formulate the MG approach.

Let $\bm{\overline{u}}_{h}$ denote the approximate layer state to the true solution $\bm{u}_{h}$, and let $\bm{U}_{h} = \left(\bm{u}^0_{h}, \ldots, \bm{u}^{N-1}_{h} \right)$ denote the array of layer states in the neural network composed of $N$ residual blocks on the \textit{level} with step size $h$. The forward propagation algorithm corresponding to \eqref{EQ:resnet} and \eqref{EQ:continuous} can be written as:
\begin{equation}
\underbrace{
   \begin{bmatrix}
       \bm{u}^0_{h} \\
       \bm{u}^1_{h} - hF(\bm{u}^0_{h},\bm{\theta}^0_{h}) \\
       \vdots \\
       \bm{u}^{N-1}_{h} - hF(\bm{u}^{N-2}_{h},\bm{\theta}^{N-2}_{h})   
   \end{bmatrix}
}_{\bm{L}_{h}\left(\bm{U}_{h},\bm{\theta}_{h}\right)}
=
\underbrace{
   \begin{bmatrix}
       \bm{F_{\text{in}}} \bm{y} \\
	   0 \\
       \vdots \\
       0
   \end{bmatrix}
}_{\bm{f}_{h}} \label{EQ:fine_level}
\end{equation}
where $\bm{F_{\text{in}}}$ corresponds to the network operator and $\bm{y}$ representing the data sample. 
Let $\overline{\bm{U}}_{h}$ denote the approximation of $\bm{U}_{h}$ such that $\bm{U}_{h} = \overline{\bm{U}}_{h} + \delta \bm{U}_{h}$ and define the notation $\bm{L}_{h}\left(\overline{\bm{U}},\bm{\theta}\right) := \bm{L}\left(\overline{\bm{U}}_{h},\bm{\theta}_{h}\right)$.
We express the solution error via the residual equation corresponding the level with step size $h$:
\begin{IEEEeqnarray}{rCl}
\bm{R}_{h} &:=& \bm{f}_{h} - \bm{L}_{h}\left(\overline{\bm{U}},\bm{\theta}\right)  \label{EQ:residual_def} \\ 
            &=& \bm{L}_{h}\left(\bm{U},\bm{\theta}\right) - \bm{L}_{h}\left(\overline{\bm{U}},\bm{\theta}\right) \\
            &=& \bm{L}_{h}\left(\overline{\bm{U}} + \delta \bm{U},\bm{\theta}\right) - \bm{L}_{h}\left(\overline{\bm{U}},\bm{\theta}\right) \label{EQ:DL_resid}
\end{IEEEeqnarray}
Equation \eqref{EQ:DL_resid} is rearranged and a variable substitution is applied such that the correction equation is given as follows:
\begin{equation}
\bm{L}_{h}\left(\bm{V},\bm{\theta}\right) = \bm{L}_{h}\left(\overline{\bm{U}},\bm{\theta}\right) + \bm{R}_{h} \label{EQ:FAS_DL}
\end{equation}
where $\bm{V} = \overline{\bm{U}} + \delta \bm{U}$. The residual equation in \eqref{EQ:FAS_DL} can be solved on a coarse level with step size $H$  for $\bm{V}$ resulting in the solution correction $\delta \bm{U} = \bm{V} - \overline{\bm{U}}$.

To construct the multilevel hierarchy of the neural network, we define a coarsening factor $c$ such that every $c^{th}$ layer within the level is additionally assigned as a member to the coarser level as demonstrated in Figure~\ref{fig:coarsening-layers}. The coarsening process can be applied repeatedly to make successive coarser levels containing fewer layers. To solve \eqref{EQ:FAS_DL} on the coarse level $H$ for Residual Networks, the fine level residual $\bm{R}_h$, approximate solution $\overline{U}_h$, and network parameters $\bm{\theta}_h$ are restricted to the coarse level by injection (copy) for every $c^{th}$-layer. 
These are denoted as C-Layers as displayed in Figure~\ref{fig:coarsening-layers}. That is, the restricted solution on the coarse level is given as follows:
\begin{equation}
\overline{\bm{U}}_H = \left(\overline{u}_H^0, \overline{u}_H^1, \ldots, \overline{u}_H^{N_H-1}\right), \ \ \text{where} \ \overline{u}_H^n = \overline{u}_h^{nc}
\end{equation}
The coarse level analogous expression of \eqref{EQ:FAS_DL} reads
\begin{equation}
\bm{L}_{H}\left(\bm{V}_H,\bm{\theta}_H\right) = \bm{L}_{H}\left(\overline{\bm{U}}_H,\bm{\theta}_H\right) + \bm{R}_{H} =: \bm{S}_{H} \label{EQ:FAS_DL_Coarse}
\end{equation}
and we define the coarse-level nonlinear operator $\bm{L}_{H}$ as
\begin{align}
\bm{L}_{H}\left(\bm{U}_{H},\bm{\theta}_{H}\right) = 
   \begin{bmatrix}
       \bm{u}^0_{H} \\
       \bm{u}^1_{H} - HF(\bm{u}^0_{H},\bm{\theta}^0_{H}) \\
       \vdots \\
       \bm{u}^{N_H-1}_{H} - HF(\bm{u}^{N_H-2}_{H},\bm{\theta}^{N_H-2}_{H})   
   \end{bmatrix}
\end{align}
where the coarse level feature-transformation operator $F_H$ is scaled by $H = ch$. This is used to approximate the information transfer between C-Layers. In a two-level MG method, $\bm{V}_H$ is solved directly in \eqref{EQ:FAS_DL_Coarse} providing the coarse-level error $\delta \bm{U}_H = \bm{V}_H - \overline{\bm{U}}_H$. The fine-level solutions in the C-Layers are updated via $\overline{u}_h^{nc} \leftarrow \overline{u}_h^{nc} + \delta \bm{u}_H^{n}$. 

\begin{figure*}[t]
\centerline{\includegraphics[width=0.7\linewidth]{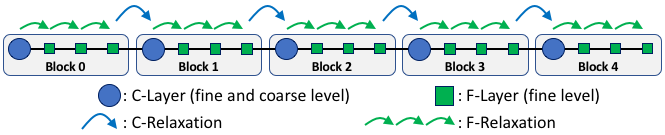}}
\caption{C-relaxation: forward propagation from preceding F-Layer to C-Layer. F-relaxation: forward propagation from C-Layer to F-Layers in block.}
\label{fig:relaxation}
\end{figure*}

Lastly, as shown in Figure~\ref{fig:relaxation}, the F-Layers within a layer block are updated directly via F-relaxation using sequential forward propagation. The transfer of information between layer blocks and partitions is achieved by C-relaxation via sequential forward propagation using the last F-Layer in the preceding layer block.

\begin{algorithm}[t]
\SetAlgoLined
1. FCF-Relaxation to $\bm{L}_{h}\left(\bm{U}_{h},\bm{\theta}_{h}\right) = f_h$\\
2. Restrict residual $\bm{R}_h$ and approximate solution $\overline{\bm{U}}_h$: \\
\ \ \ $\bm{R}_H^n \leftarrow \bm{R}_h^{nc}$, $\overline{\bm{U}}_H^n \leftarrow \overline{\bm{U}}_h^{nc}$ \ \ for $n = 0, \ldots, N_H-1$\\
3. Solve $\bm{L}_{H}\left(\bm{V}_H,\bm{\theta}_H\right) = \bm{L}_{H}\left(\overline{\bm{U}}_H,\bm{\theta}_H\right) + \bm{R}_{H}$\\
4. Solve coarse-level error: $\delta \bm{U}_H = \bm{V}_H - \overline{\bm{U}}_H$\\
5. Update fine-level approximate solution: \\
\ \ \ $\overline{u}_h^{nc} \leftarrow \overline{u}_h^{nc} + \delta \bm{u}_H^{n}$ \ \ for $n = 0, \ldots, N_H-1$\\
6. Check convergence:\\
\If{$||\bm{R}_h|| \leq tol$}{Return;}{Proceed to step 1;}
\caption{ResNet Multigrid FAS Scheme}
\label{Alg1}
\end{algorithm}

\subsection{Implementation}
Algorithm \ref{Alg1} outlines the solution procedure using the MG FAS scheme for a two-level network approach. The implementation of the algorithm for forward-propagation was performed in C++ wrapping NVIDIA's CuDNN \cite{Chetlur:2014} library kernels. CuDNN's API allows the use of CUDA Streams enabling asynchronous and concurrent execution of kernels. Each layer block shown in Figure~\ref{fig:relaxation} is assigned a unique CUDA Stream allowing threads blocks to be solved concurrently. To achieve concurrent execution, kernels are launched in parallel using OpenMP assigning one CPU thread per layer block. Additionally, the usage of multiple GPUs for model-parallelism is implemented using the Message Passing Interface (MPI) \cite{Gropp:1999}. The layer blocks are distributed into contiguous model partitions across GPUs, and distributed memory communication is performed during the C-relaxation phase for abutted layer blocks residing on different GPUs.

\begin{figure}[ht]
\centerline{\includegraphics[scale=0.29]{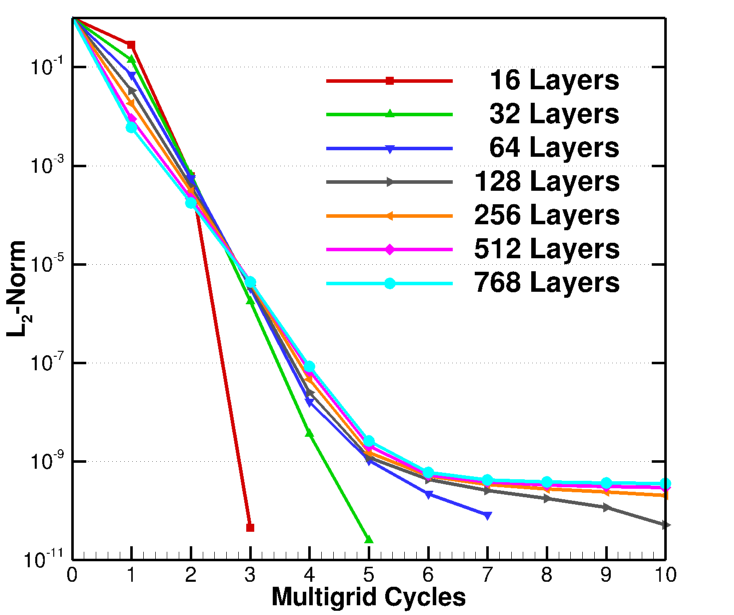}}
\caption{Residual convergence rates for varying network layer depths demonstrating network layer-size independence.}
\label{fig:convergence}
\end{figure}

\begin{figure*}[ht]
\centerline{\includegraphics[scale=0.5]{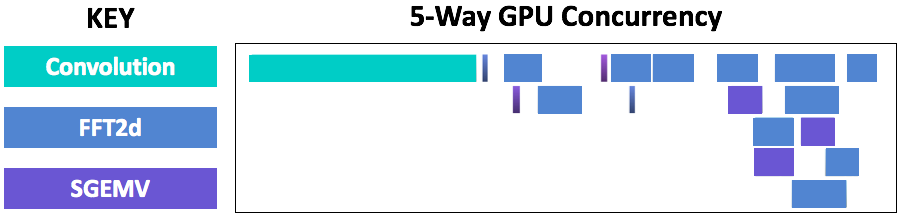}}
\caption{Compute execution time line within one NVIDIA V100 GPU demonstrating kernel concurrency.}
\label{fig:gpu}
\end{figure*}

\section{Numerical Experiments}
Herein, we demonstrate the algorithm using the MIT Supercloud supercomputer TX-GAIA \cite{Reuther:2018}, which is composed of 448 compute nodes, each node contains two 2.5GHz Intel Xeon Gold 6248 (20-core) processors with 384 GB of RAM. Every node contains two NVIDIA Tesla 32GB V100 GPUs, both connected to the first CPU only. These results utilize a 25 Gb/s Ethernet interconnect using Mellanox ConnectX-5 adapters connected to a single non-blocking Arista DCS-7516 Ethernet core switch, noting that NVLink is not present in the system. Note that Multigrid is independent of architecture, and can be employed on legacy hardware such as CPUs or FPGAs. MG can be used in conjunction with data-parallelism for the training of neural networks, but we restrict our experiments to exclusively explore performance metrics of the algorithm as a model-parallelism approach. We utilize the MNIST \cite{LeCun:1998} data set for testing, which is composed of 60,000 hand-written digits encoded as 28 $\times$ 28 pixel single channel (grey-scale) images.

\subsection{Verification and Layer-Independent Convergence}
To ensure algorithmic correctness, we perform a convergence study on residual networks of various layer-depth counts. 
Figure~\ref{fig:convergence} demonstrates the $L_2$-norm convergence histories of the residual vector $\bm{R}_h$ found in \eqref{EQ:residual_def}. Note that as more Multigrid Cycles are utilized, $\left\lVert \bm{R}_h \right\rVert_{L_2} \rightarrow 0$ implies $\overline{\bm{U}} \rightarrow \bm{U}$ meaning the approximate solution converges to the analytic solution normally obtained by sequential forward propagation.

Further, Figure~\ref{fig:convergence} shows the norm values for all networks composed of different layer counts converge with similar behavior down to $\left\lVert \bm{R}_h \right\rVert_{L_2} \leq 10^{-9}$. This indicates an important property that the solution convergence is independent of the network depth. Thus, given enough computational resources, this property will provide approximate constant time to solution evaluation of arbitrarily-deep networks. 

Secondarily, an approach known as \textit{early stopping} from \textit{One-Shot} methods \cite{Bosse:2014} can be applied such that the solution convergence is stopped after a fixed number of iterations. This provides an approximate solution which can be used to generate approximate neural network gradients for the training optimization process. For this work, we find that \textbf{two} cycles suffice, providing accurate state estimates resulting in approximately the same Top-1 error rates after each epoch of training for the MNIST data set.

\subsection{GPU Concurrency}
The aforementioned implementation utilizes NVIDIA's Cuda Streams to enable concurrent kernel execution. Using NVIDIA's Profiler (nvprof), Figure~\ref{fig:gpu} shows an excerpt of the kernel execution time line for a single GPU computing a MG cycle composed of convolution layers. As seen in the figure, 5-way kernel concurrency is achieved. However, the number of registers within the GPU prevents multiple convolution kernels from executing simultaneously, thus limiting the performance for the exposed parallelism. At present, the concurrent execution capability is not fully realized due to this limitation but will allow for higher throughput in future architectures.

\subsection{Multi-GPU Parallel Inference}
Next, we test single image inference via the Multigrid Layer-Parallelism approach on a 4,096 layer Residual Neural Network containing 3,248,524 parameters. 
The network is composed of an opening convolution layer with a $7\times 7$ kernel, 50 output channels, one layer of padding, single striding, and a Rectified Linear Unit (ReLU) activation function. This layer is followed by 4,092 residual convolution layers of the same dimensions and activation functions, and terminated by a fully connected layer mapping the output to a 10-element softmax function paired with cross-entropy loss.

\begin{figure*} 
  \centering
  \vspace{-1\baselineskip}
  \subfloat[\label{fig:inference} Inference Performance]{\includegraphics[width=0.325\linewidth]{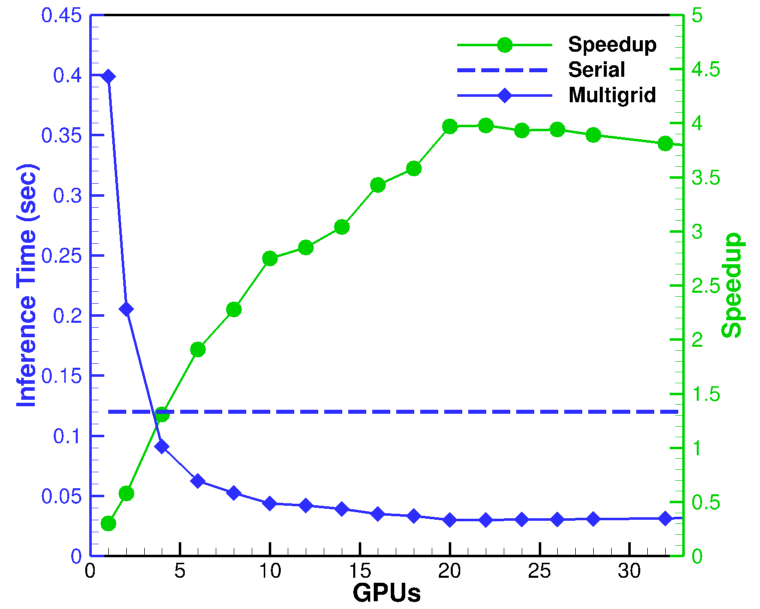}}
  \subfloat[\label{fig:4096Layers:b} Training Performance]{\includegraphics[width=0.315\linewidth]{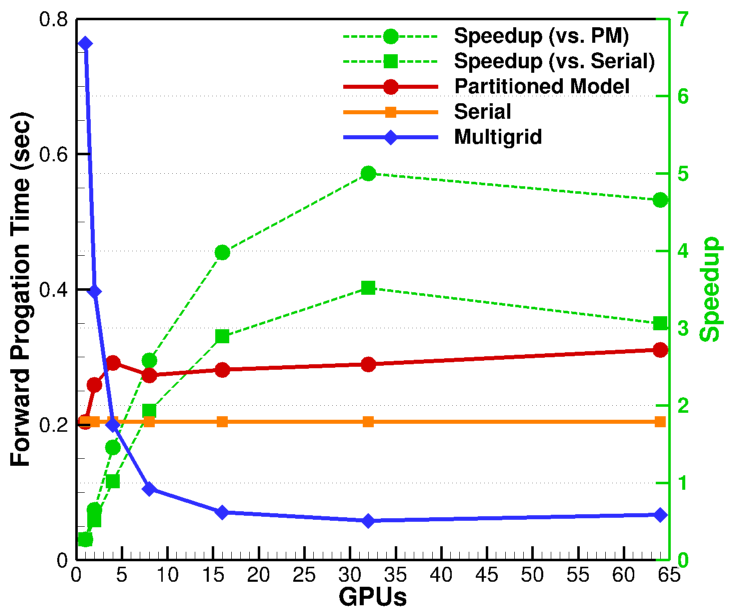}}
  \subfloat[\label{fig:4096Layers:c} Timing Decomposition]{\includegraphics[width=0.298\linewidth]{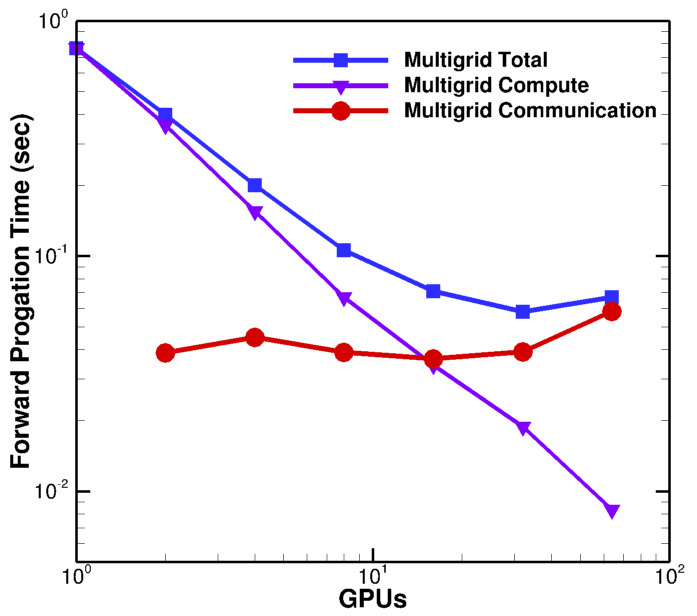}}
  \caption{Parallel strong scaling of forward propagation for a 4,096-layer Residual Neural Network composed of 3,248,534 parameters.}
  \vspace{-1\baselineskip}
  \label{fig:4096Layers}
\end{figure*}

Figure~\ref{fig:inference} shows the performance profile comparison of the Multigrid algorithm compared to the serial forward propagation approach. When using a single GPU, Multigrid is approximately four times slower; this is expected as MG is an iterative method requiring multiple evaluations of the network. However, as we distribute the model over multiple GPUs, particularly greater than three GPUs, the inference time is reduced achieving a speedup over the sequential approach. Using four GPUs, MG is 1.25x faster and when using 24 GPUs, it is 4x faster. We note that as the model is partitioned more finely, more distributed memory communication is introduced causing the performance to be limited. This is particularly true for heterogeneous architectures composed of CPUs as hosts and GPUs as accelerators.

\subsection{Multi-GPU Parallel Training: 3 Million Parameters}
To study the effect of communication, we examine the training phase of the same neural network used for the inference study.
Figure~\ref{fig:4096Layers:b} shows the performance comparisons of the serial method, Multigrid, and the traditional sequential partitioned model (PM) layer-parallel method\footnote{The model is partitioned, layer-wise, over multiple GPUs; each GPU contains a subset of the network's layers, and the evaluation of the network is serialized across the GPUs.}. When four or more GPUs are utilized, MG is the superior algorithm providing up to 3.5x speedup over serial and 5x speedup over the PM method. 
Figure~\ref{fig:4096Layers:c} demonstrates the timing decomposition elucidating the effect of communication on parallel strong scaling. Computation is evenly distributed across each compute device as more devices are added, and the parallel execution in log-log scale shows near-perfect parallel strong scaling. In contrast, as more GPUs are used to partition the network, the communication becomes the performance bottleneck. In the case of 64 GPUs, 97\% of the evaluation time is consumed by the communication.

\begin{figure}[ht]
\centerline{\includegraphics[scale=0.31]{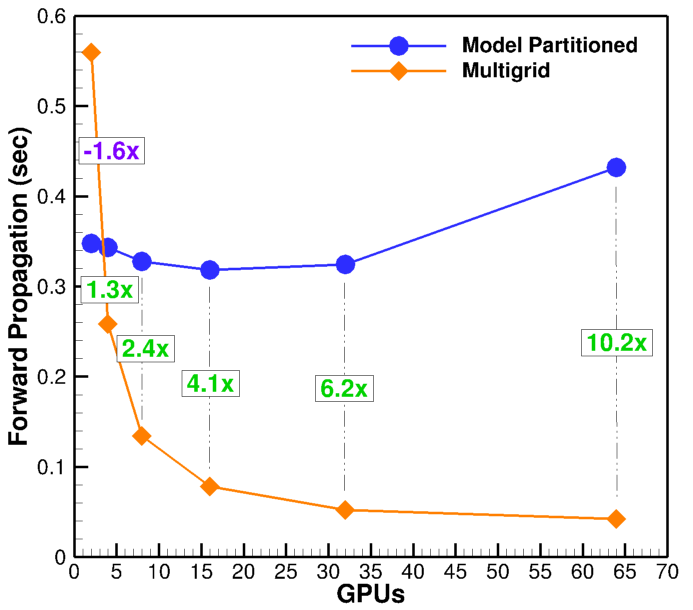}}
\caption{4,115-layer Residual Network composed of 2,071,328,150 parameters. Model Partitioned is serial execution distributed across multiple devices.}
\label{fig:massive}
\end{figure}

\subsection{Multi-GPU Parallel Training: 2 Billion Parameters}
\vspace{0.25cm}
The 3.25 million parameter model highlighted the effects of the ratio of computation to communication with respect to the achievable performance via Multigrid Layer-Parallelism. To exploit this further, we derive a computation-intensive residual network composed of 4,115 layers: an opening convolution layer [$7\times 7$ kernel, 20 output channels, padding=1, single stride, ReLU], 16 repeated sequence blocks containing: one fully connected layer with matching input and output dimensions, and 256 residual convolution layers matching the opening convolution layer specifications. Lastly, the network is computed with a fully connected layer to a softmax function with cross-entropy loss. This model contains a total of \textbf{2,071,328,150} parameters, which is on the order of the Megatron-LM \cite{Shoeybi:2019} model containing 8.3 billion parameters.

The 18 fully connected layers greatly increase the floating-point operation (FLOP) counts, therefore increasing the arithmetic complexity defined as the ratio of FLOPs performed to the bytes transferred for the computation. Additionally, the number of model parameters requires some form of model parallelism as it is too large to reside on a single GPU. Thus, we will compare Multigrid to the traditional \textit{Layer-Wise Parallelism}, which partitions the model across compute devices and performs serial network evaluation.

Figure~\ref{fig:massive} displays the parallel strong scaling performance comparing the Multigrid algorithm to the layer-wise parallelism method, denoted as Model Partitioned. 
While each method utilizes the same number of GPUs, it is clear that Multigrid outperforms its counterpart when using at least four GPUs. 
A speedup of 1.3x is achieved when four GPUs are utilized and a speedup of 10.2x is achieved when 64 GPUs are utilized. In the case of using four GPUs for MG, the computation to communication ratio is 92.8\%, highlighting experimentally, the high arithmetic intensity. This ratio drops to 34.5\% at 64 GPUs implying the performance becomes limited due to communication exhibiting low arithmetic intensity.

\section{Conclusion}
This work exposed more parallelism within supervised Deep Learning, which can be paired with data-parallelism approaches. For model training and inference, Multigrid demonstrates avenues of increased throughput for compatible network architectures  exhibiting high arithmetic intensity relative to the computing architecture utilized. We demonstrated performance gains up to 10.2x using NVIDIA V100 GPUs, a heterogeneous architecture tailored for high-compute and low-data movement. The algorithm is favorable towards architectures with improved communication interconnects between devices.

Lastly, the current state of hardware demonstrates limited kernel-execution concurrency, specifically in the context of convolution operators. Thus, the performance of this algorithm and implementation has not been fully realized. This algorithm will improve in future computer architectures containing massive thread-based hardware parallelism indicative of future exascale-era architectures\cite{Messina:2017}. Additionally, this algorithm can be implemented in conjunction with data-parallel techniques for multiplicative-compounding parallelism as required for next-generation supercomputing systems.

\section*{Acknowledgment}
The authors acknowledge The authors wish to acknowledge the following individuals for their contributions and support: Dave Martinez, Steve Rejto, Marc Zissman along with William Arcand,
David Bestor, William Bergeron, Chansup Byun, Matthew
Hubbell, Michael Houle, Anna Klein, Peter Michaleas, Lauren Miliechin, Julie
Mullen, Andrew Prout, Antonio Rosa, and Charles Yee.

\end{document}